\documentclass{article}

\usepackage{PRIMEarxiv}

\usepackage[utf8]{inputenc} 
\usepackage[T1]{fontenc}    
\usepackage{hyperref}       
\usepackage{url}            
\usepackage{booktabs}       
\usepackage{amsfonts}       
\usepackage{nicefrac}       
\usepackage{microtype}      
\usepackage{lipsum}
\usepackage{fancyhdr}       
\usepackage{graphicx}       
\graphicspath{{media/}}     
\usepackage{amsmath}
\usepackage{makecell}

\title{Improving Apple Object Detection with Occlusion-Enhanced Distillation
}

\author{
  {Yuanping Shi$^{1,2,3}$, Yanheng Ma$^{1}$, Liang Geng$^{2,3}$, Lina Chu$^{1}$, Bingxuan Li$^{1}$} \\
  1 Department of UAV Engineering, Shijiazhuang Campus, Army Engineering University, Shijiazhuang, 050003, China \\
  2 College of Mechanical and Electrical Engineering, Shijiazhuang University, Shijiazhuang, 050035, China \\
  3 Shijiazhuang Key Laboratory of Agricultural Robotics Intelligent Perception, Shijiazhuang, 050035, China \\
  \texttt{1102061@sjzc.edu.cn (Y.Shi); mamyh11@163.com (Y.Ma); liang\_geng@bupt.edu.cn (L.Geng);} \\
  \texttt{chulina@aeu.edu.cn (L.Chu); libingxuan2021@aeu.edu.cn (B.Li)} \\
}

\begin{document}
\maketitle

\begin{abstract}
Apples growing in natural environments often face severe visual obstructions from leaves and branches. This significantly increases the risk of false detections in object detection tasks, thereby escalating the challenge. Addressing this issue, we introduce a technique called "Occlusion-Enhanced Distillation" (OED). This approach utilizes occlusion information to regularize the learning of semantically aligned features on occluded datasets and employs Exponential Moving Average (EMA) to enhance training stability. Specifically, we first design an occlusion-enhanced dataset that integrates Grounding DINO and SAM methods to extract occluding elements such as leaves and branches from each sample, creating occlusion examples that reflect the natural growth state of fruits. Additionally, we propose a multi-scale knowledge distillation strategy, where the student network uses images with increased occlusions as inputs, while the teacher network employs images without natural occlusions. Through this setup, the strategy guides the student network to learn from the teacher across scales of semantic and local features alignment, effectively narrowing the feature distance between occluded and non-occluded targets and enhancing the robustness of object detection. Lastly, to improve the stability of the student network, we introduce the EMA strategy, which aids the student network in learning more generalized feature expressions that are less affected by the noise of individual image occlusions. Our method significantly outperforms current state-of-the-art techniques through extensive comparative experiments.
\end{abstract}


\section{Introduction}
In modern agricultural automation, detecting fruits within orchards is a crucial task \cite{1, 2, 3}. These tasks significantly assist farmers in managing resources, optimizing production processes, and making data-supported decisions during the harvest period. Particularly in the field of mechanized harvesting, efficient robotic systems are required to accurately identify fruits amidst cluttered canopies, which not only substantially enhances picking efficiency but also greatly reduces the labor intensity and cost associated with manual harvesting.

\begin{figure}[htbp]
	\includegraphics[width=\linewidth]{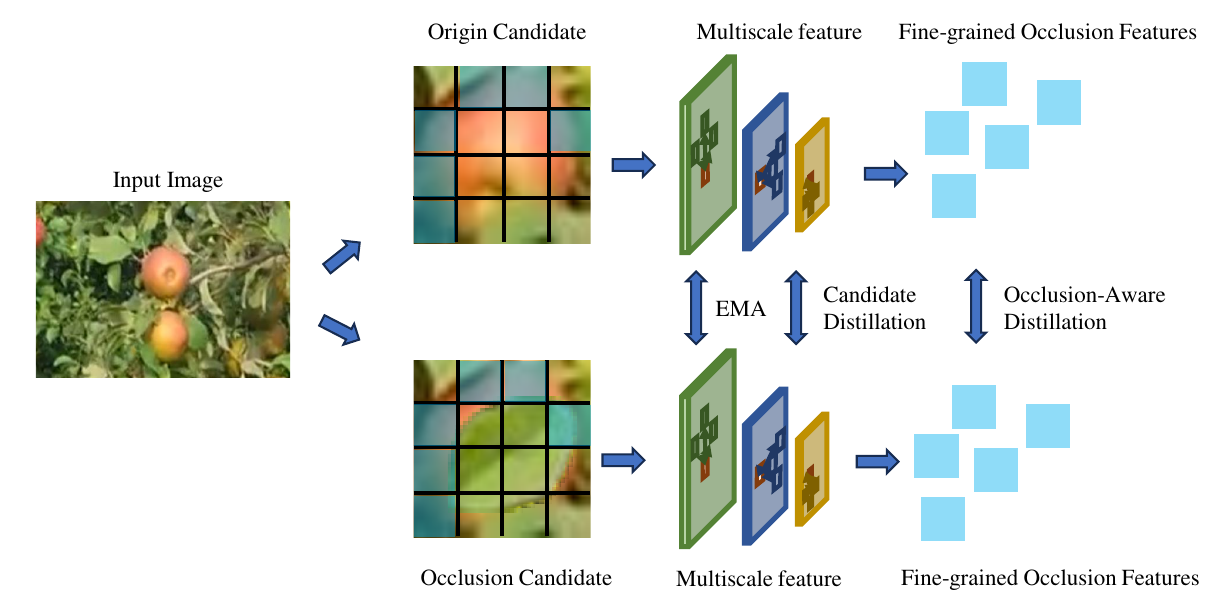}
	\caption{Method Overview: We employ Multi-scale Feature Distillation to address the training challenges posed by substantial morphological differences in targets under severe occlusion. Specifically, Multi-scale Feature Distillation is divided into two parts: Candidate Distillation and Occlusion-Aware Distillation. Through weighted processes, relevant information is distilled from multi-scale features for knowledge transfer. Additionally, to enhance training stability, we utilize exponential moving averages.\label{fig01}}
\end{figure}

However, occlusion is a common occurrence in fruit trees growing under natural conditions. Leaves and branches in the images may be very close to the position of the fruits, or even appear to merge with them, often leading to erroneous identification of fruits \cite{4}. In some cases, the occluded parts contain only limited target information, leading to significant variations in the appearance of objects under different occlusion states. As shown in Figure \ref{fig01}, we simulate natural conditions by adding random occlusions to the original images, significantly altering the shape of the apple at the center due to occluding leaves. Under these circumstances, traditional object detection frameworks struggle to adapt to the learning needs of heavily occluded samples. This is primarily because convolution-based methods like Fast R-CNN \cite{11}, Faster R-CNN \cite{12}, Mask R-CNN \cite{13}, YOLO \cite{14}, and SSD \cite{17}, as well as Transformer-based algorithms such as Deformable DETR \cite{72} and DINO \cite{73}, require direct mapping of fruit features to labels to perform object detection. However, due to the typically limited scale and complex occlusion scenarios of agricultural datasets, these methods often fail to effectively learn the significant shape variations caused by occlusions. Addressing these challenges necessitates more refined algorithmic design.

To address the issue of strong occlusions, traditional methods often rely on forcibly extracting image features and aligning them in semantic space. However, the significant morphological differences before and after occlusion lead to instability during the training process. Inspired by how humans infer the complete object through partial unoccluded information, this paper proposes a strategy that diverges from previous approaches. Instead of pursuing alignment of semantic features at the target scale, we utilize the similarities in the unoccluded parts of the target to guide the learning process. For instance, in the Occlusion Image of Figure \ref{fig01}, the area marked by the blue square shows the unoccluded part of the apple. Under the framework based on Deformable DETR \cite{72}, we select these features and align them across multiple scales, effectively mitigating the learning instability caused by morphological differences at the target scale.

Specifically, we propose a new framework named Occlusion-Enhanced Distillation (OED). OED utilizes knowledge distillation techniques to align the feature embeddings of fruits before and after occlusion, thereby enhancing robustness against random occlusions. Our method is divided into three parts: Firstly, Grounding DINO \cite{52} with the text prompt 'leaves' is applied on the dataset to obtain bounding boxes of the occluded objects, which helps capture the occlusion characteristics of fruits in their natural growing states, ensuring consistency with natural growth and lighting conditions. Next, we use SAM \cite{42} to generate precise occlusion masks within these bounding boxes. Based on the annotated fruit masks in the dataset, we randomly occlude the fruits. Between the teacher and student models, we learn to align the feature embeddings of occluded and unoccluded samples through self-distillation of knowledge. Considering the significant changes in the appearance of fruits under severe occlusion, previous methods of directly performing knowledge distillation between teacher and student networks to achieve feature alignment \cite{74, 75} may no longer be suitable. This is because significant morphological differences can lead to difficulties in the learning process. To overcome this challenge, we compute the similarity of feature responses between the student and teacher models at multiple scales and perform optimal patch matching to find semantically similar features at a specific scale. Additionally, we adopt an Exponential Moving Average (EMA) strategy for the student model's weights to minimize the impact on the student network under extreme or irregular data perturbations.

The main contributions of this study can be summarized as follows:

\begin{itemize}
	\item A novel method combining Grounding DINO with SAM was designed to construct an occlusion-enhanced dataset for teacher-student network self-distillation.
	\item A multi-scale feature similarity assessment method was developed, implementing optimal patch matching to effectively address the significant differences in appearance between occluded and unoccluded fruits while retaining similar semantic features.
	\item An Exponential Moving Average (EMA) strategy for student weights was introduced, enhancing the training stability of the student network when dealing with extreme and irregular data.
\end{itemize}

\section{Related Work}

Object detection: In the field of agricultural automation, particularly in the detection of apple targets in orchards, occlusion presents a ubiquitous and challenging problem. In real-world scenarios, apples are often obscured by leaves, branches, or other fruits, significantly increasing the difficulty of detection. Currently, object detection techniques based on Convolutional Neural Networks (CNNs) and Transformers each demonstrate distinct advantages and limitations in handling occluded target detection tasks. CNN-based methods, due to their robust feature extraction capabilities, are widely used for recognizing and locating complex objects in images, yet they may be limited in handling occlusions and overlapping objects. In contrast, Transformer-based techniques, such as DETR \cite{9}, by leveraging global contextual information, are better equipped to understand the occlusions and the relative spatial relationships between objects in images, thus enhancing the detection accuracy under occluded conditions.

Existing research primarily employs Convolutional Neural Network (CNN)-based methods for addressing object detection issues. For example, Faster-RCNN \cite{12} significantly improves processing speed and accuracy by introducing a Region Proposal Network (RPN), which generates candidate object regions and utilizes deep learning models for direct feature learning and classification. Additionally, algorithms such as YOLO \cite{14} and SSD \cite{17} approach detection tasks as a single regression problem, further enhancing processing speed. YOLO predicts bounding boxes and class probabilities directly across the entire image, markedly accelerating performance. Concurrently, SSD enhances the detection capability for apples of varying sizes by performing detection on feature maps at multiple scales. However, the efficiency and accuracy of these methods tend to be compromised when the degree of occlusion increases \cite{76}. Particularly when apples are mostly obscured by leaves, these detection models often struggle to accurately distinguish objects. This issue primarily arises because they fail to capture sufficient visual feature information in cases of severe occlusion \cite{77}. Convolutional networks rely on local or boundary information to recognize objects, but in heavily occluded scenarios, effective features may be obscured by obstructions, leading to challenges in making accurate classifications.

In recent years, Transformer models have increasingly demonstrated their unique advantages in the field of computer vision, particularly excelling in handling occluded object detection. For instance, DETR (Detection Transformer) \cite{9} and its derivative models \cite{72, 73} introduced an innovative approach by transforming the task of object detection into a set prediction problem, thus eliminating the need for complex post-processing steps typical of traditional region proposal-based detection methods. DETR utilizes the self-attention mechanism to process the global dependencies in images, which is particularly crucial for recognizing partially occluded objects against complex backgrounds. Despite their effective handling of global information through the self-attention mechanism, Transformer models may still face limitations in scenarios of extreme occlusion, where objects are mostly obscured. This is because even global attention mechanisms struggle to accurately infer complete object information from very limited visible cues, especially when occlusion is severe enough to obscure key visual features of the object.

Knowledge distillation: Distillation methods enhance the learning effectiveness of student models by utilizing the relationships between different samples. For instance, by comparing how similar objects are processed in different images, student models can better understand the general appearance of objects under complex backgrounds or occluded conditions. Although guidance from high-level features assists the student model's learning during hint-based training \cite{78}, the vast differences in target shapes under extreme occlusion may result in a lack of holistic information. The literature \cite{79} suggests that forcing student models to mimic teacher models on features specified by attention maps can mislead the focus of attention when dealing with random occlusions like trees. Moreover, \cite{80} proposes knowledge distillation using relationships between different samples, but direct similarity distillation across samples performs poorly under severe occlusion. Additionally, the distribution matching approach in \cite{81} relies on complete target information, making it difficult to perform accurately during severe occlusion. Research \cite{82} indicates that full feature imitation may lead to overfitting in the occluded parts by the student model, and has found that this approach might reduce the performance of the student model.

\begin{figure}[htbp]
	\includegraphics[width=\linewidth]{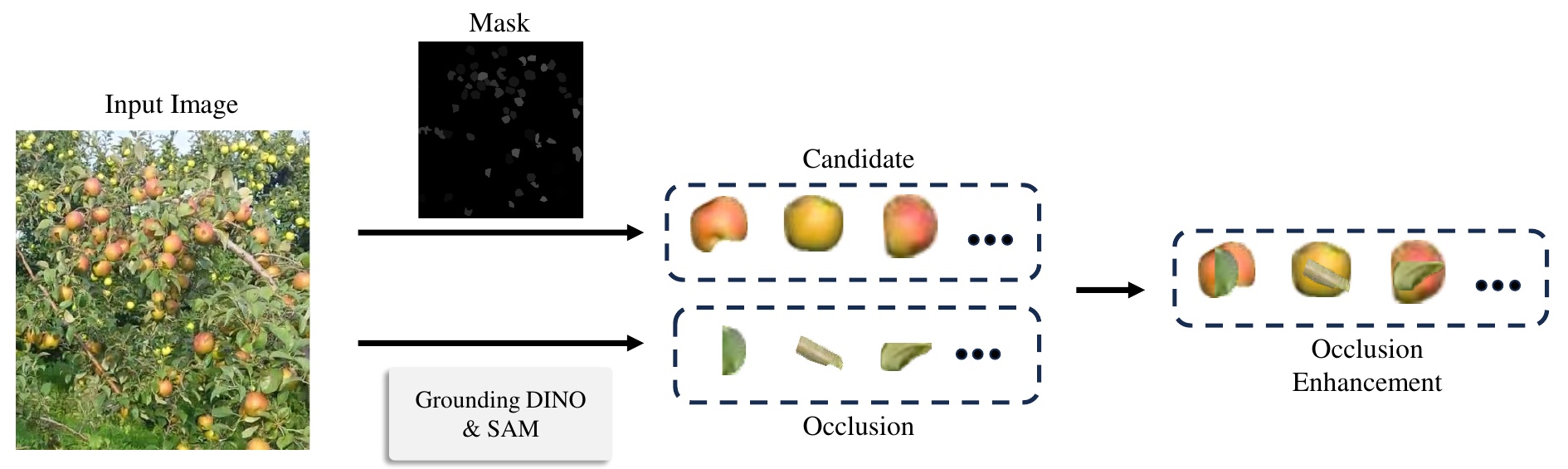}
	\caption{Data Occlusion Augmentation: Utilizing annotated masks from the dataset, occluders are extracted using Grounding Dino \cite{52} and SAM \cite{43} to occlude the targets.\label{fig02}}
\end{figure}

\section{Data Occlusion Enhancement}

To conduct distillation learning on the MinneApple dataset, we implemented a series of occlusion data augmentation operations. Initially, RGB images were segmented using annotated masks from the dataset to create a set of instance templates. Subsequently, we employed a pretrained zero-shot detector, namely Grounding Dino \cite{52}, using text prompts such as "leaves" and "branches" to identify occluders in natural scenes and obtain their initial bounding boxes. Following this, we applied SAM \cite{43} to generate masks based on these bounding boxes, thereby constructing a set of occlusion templates. Ultimately, by combining the occlusion template set with the instance template set, we built a candidate set for querying operations within the instance set.

\begin{figure}[htbp]
	\includegraphics[width=\linewidth]{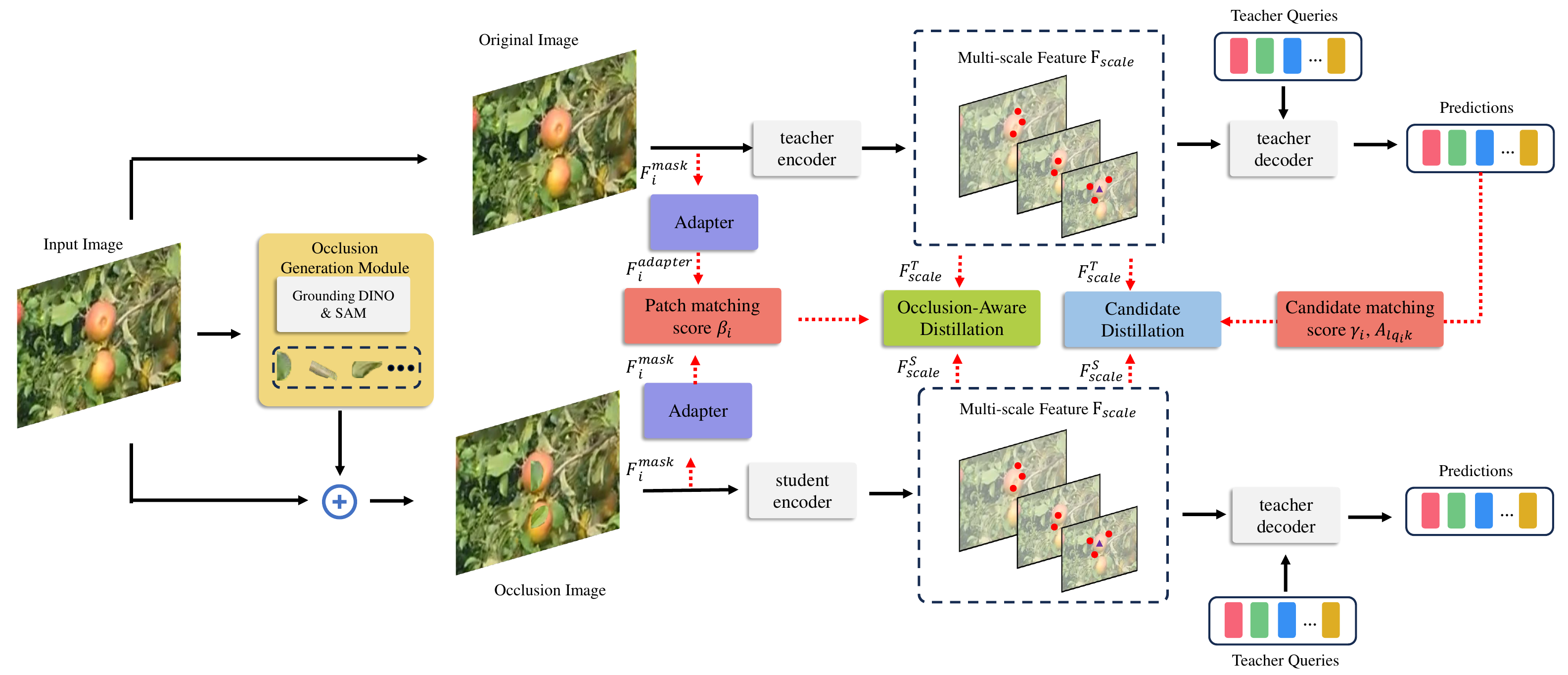}
	\caption{Model Architecture: We employ Deformable DETR \cite{72} as the backbone model to extract multi-scale features. Based on these multi-scale features, we implement two levels of knowledge distillation: Candidate Distillation and Occlusion-Aware Distillation. The feature weights used in these distillation processes are derived from the detector's query responses and the fine-grained matching of occlusions.\label{fig03}}
\end{figure}

\section{Occlusion-Enhanced Distillation (OED) Model}

Occlusion-Enhanced Distillation (OED) is a feature distillation technique specifically designed for dealing with heavy occlusion, aimed at improving the performance of deep learning models when processing occluded images. This method employs a hierarchical feature imitation strategy, meticulously aligning and optimizing the feature representation from the overall objective layer down to the local fine-grained occlusion layer. This approach is particularly suited for occlusion handling in visual recognition tasks, significantly enhancing model robustness and accuracy in complex environments (Section 4.1). Additionally, to boost model stability and convergence during training, we utilize an Exponential Moving Average (EMA) strategy. This method reduces variance during training by weighting and smoothing model parameters, giving greater weight to the most recent observations, thereby enhancing the model's generalization capability in occlusion handling tasks (Section 4.2).

\subsection{Multi-Scale Feature Distillation}

Multi-scale feature extraction techniques significantly enhance the ability to capture target information at various levels within an image, from macroscopic instance levels to microscopic fine-grained levels, providing support for in-depth analysis. This study employs Deformable DETR \cite{72} as the backbone model, optimizing multi-scale feature extraction through its unique deformable attention mechanism. Given the significant changes in the appearance of targets under occlusion conditions, this research designs a dual-level knowledge distillation framework, focusing on Candidate Distillation and Occlusion-Aware Distillation. Candidate Distillation is obtained through the target detection head of the backbone, while the occlusion-aware feature level focuses on extracting features from prominent unoccluded parts within the apple detection frame, utilizing adapters to extract from embedding features prior to the backbone encoder. These refined feature informations interact with multi-scale feature maps, and knowledge is distilled through the computed weights to enhance the model's ability to recognize targets in complex environments and its robustness.

Multi-scale Feature Extraction: One of the common challenges faced in detecting apples on fruit trees is the issue of small targets. Existing research \cite{85} indicates that extracting features from multi-scale feature maps is particularly beneficial for small target detection. Based on this consensus, we employ Deformable DETR \cite{72} as the backbone of the model, utilizing its optimized capabilities for multi-scale processing. The set of multi-scale feature maps input is defined as $\{x_l\}_{l=1}^L$, where each feature map $x_l \in R^{C \times H_l \times W_l}$. The normalized coordinates of the reference point for the query element $q$ are denoted as $\hat{p}_q \in [0,1]^2$. Multi-scale features $F_{scale}$ are defined by the following formula:

\begin{equation}
	\mathrm{F}_{scale}(\mathrm{q})=\sum_{l=1}^L \sum_{k=1}^K A_{l q k} \cdot W \cdot x_l\left(\varphi_l\left(\hat{p}_q\right)+\Delta p_{l q k}\right),
\end{equation}

Here, $l$ denotes the feature level, and $k$ represents the sampling point. The sampling offset $\Delta p_{lqk}$ and the attention weight $A_{lqk}$ are defined as the offset and weight for the $k-th$ sampling point in the lll-th feature level, respectively. All attention weights satisfy the normalization condition $\sum_{l=1}^L \sum_{k=1}^K A_{lqk} = 1$. The function $\phi_l(\hat{p}_q)$ rescales the normalized coordinates of the query point $\hat{p}_q$ to the corresponding coordinate location on the input feature map of the $l-th$ level.

Candidate Distillation: In the task of apple detection, occlusion significantly impairs detection performance, mainly due to the target morphological differences caused by occlusion, which in turn affects the feature representation within the network. Therefore, refining the knowledge of the teacher model at the feature level is crucial for enhancing the imitation capabilities of the student network. To enable the student model to effectively mimic the spatial features of the teacher model, we have defined the following knowledge distillation objective function:

\begin{equation}
	\mathcal{L}_{\mathrm{f}}=\left\|F_{scale}^T-F_{scale}^S\right\|_2^2,
\end{equation}

Here, $F_{scale}^T \in R^{H \times W \times d}$ and $F_{scale}^S \in R^{H \times W \times d}$ represent the feature representations generated by the teacher and student models, respectively. $H$ and $W$ denote the height and width of the features, and $d$ is the number of channels in the features of both the teacher and student models. However, in the object detection framework based on DETR \cite{9}, the detection head assigns a probability score for each feature patch, which may lead to the inclusion of many features unrelated to the targets in the objective function. Therefore, we selectively use the features from the teacher network by considering the Intersection Over Union (IOU) between the teacher's predicted scores $c_i$ and bounding boxes $b_i$, to enhance the relevance of target predictions and more effectively reduce the distance between the feature representations of the teacher and student models. Our designed method for computing weights is as follows:

\begin{equation}
	S_{c_i}=\frac{\exp \left(c_i\right)}{\sum_{j=1}^N \exp \left(c_j\right)},
\end{equation}

\begin{equation}
	S_{b_i}=\frac{\exp \left(b_i\right)}{\sum_{j=1}^N \exp \left(b_j\right)},
\end{equation}

\begin{equation}
	\gamma_i = S_{c_i} * S_{b_i},
\end{equation}

Here, $S_{c_i}$ and $S_{b_i}$ are the softmax weights for the classification scores and bounding box IOU scores, respectively, and the candidate weight coefficient $\gamma_i$ is the product of the two. $N$ is the total number of target predictions. Using softmax transforms the classification scores and IOU scores into a smoother probability distribution, which helps the model to more evenly consider the contributions of all queries. The final candidate feature distillation loss function is:

\begin{equation}
	\mathcal{L}_{candidate}=\sum_{i=1}^N \gamma_i \cdot A_{l q_i k} \cdot\left\|F_{scale}^T-F_{scale}^S\right\|_2^2.
\end{equation}

\begin{figure}[htbp]
	\centering
	\includegraphics[width=0.25\textwidth]{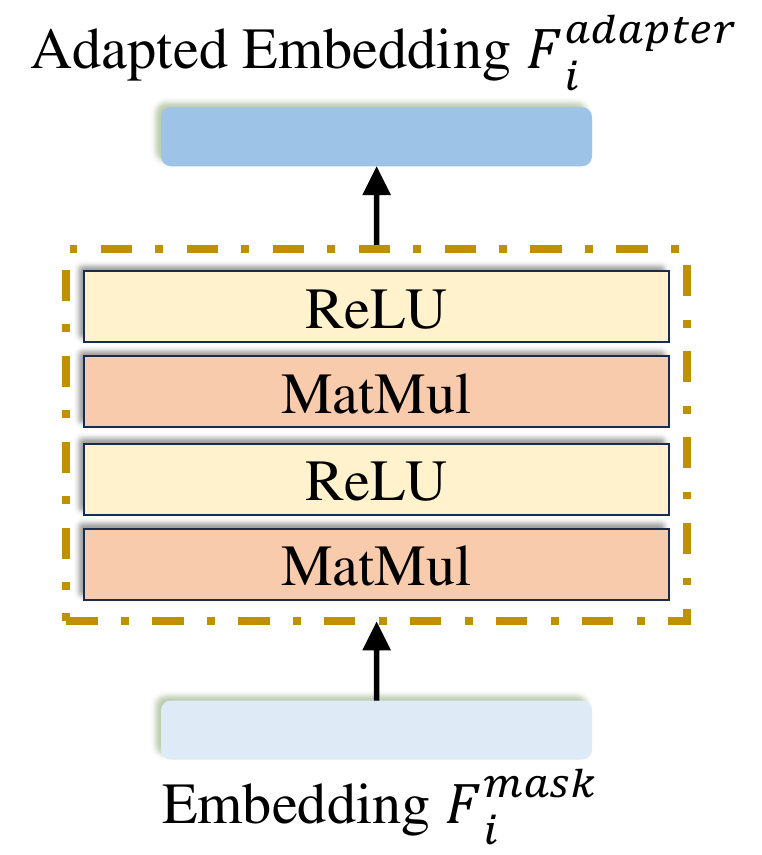}
	\caption{The adapter employs a Vision Transformer (ViT) \cite{86} structure consisting of two layers of self-attention mechanisms, specifically designed for further feature embedding.\label{fig04}}
\end{figure}

Occlusion-Aware Distillation: In the case of occluded fruits, the appearance of the fruit regions may exhibit significant differences. Therefore, relying solely on traditional global feature-based knowledge distillation methods (as shown in Equation (7)) may result in suboptimal learning performance due to substantial morphological differences. To achieve more effective feature distillation, this study focuses on comparing feature similarities between the student model and the teacher model at multiple scales, with particular attention to fine-grained knowledge distillation of the unoccluded parts of the target. This approach mimics the human mechanism of processing occluded objects by inferring information about the entire target through analyzing the visible parts. By employing this strategy, we can more accurately identify and utilize the key features of the unoccluded parts, thereby achieving more effective knowledge transfer and model training.

First, the bounding box coordinates $\{(x_{min}, y_{min}, x_{max}, y_{max})_i\}_{i=1}^M$ defined in the RGB image are converted to the corresponding feature map coordinates of the Deformable DETR \cite{72} convolutional layers, where $M$ is the number of annotated targets in the image. This conversion process considers the stride and padding of each layer, making appropriate adjustments to the position and size of the bounding boxes. Each coordinate point of the bounding box is scaled according to the stride of each layer and adjusted according to the padding. With these converted coordinates, the regions covered by the bounding boxes on the feature map can be identified, which are directly related to the activated convolutional kernel regions $\{F_i^{mask}\}_{i=1}^M$. The mask features $F_i^{mask}$ for each target are processed through the adapter to generate enhanced feature maps:

\begin{equation}
	F_i^{adapter} = Adapter(F_i^{mask})
\end{equation}

The adapter here employs a Vision Transformer (ViT) \cite{86} structure consisting of two layers of self-attention mechanisms, specifically designed for further feature embedding, as illustrated in Figure. \ref{fig04}. After processing by the adapter, we extract features from both the teacher and student models, thus obtaining the features $f_{i,j}^{T\_adapter, patch}$ and $f_{i,k}^{S\_adapter, patch}$ for each patch j and $k$ in each detection box $i$. These features are used to construct a similarity matrix $S$, where the similarity $S_{j,k}$ is calculated using cosine similarity:

\begin{equation}
	s_{j, k}={CosineSimilarity}\left(f_{i, j}^{T\_adapter, patch}, f_{i, k}^{S\_adapter, patch}\right)
\end{equation}

Next, we search for the patch with the highest similarity, which is accomplished through the following formula:

\begin{equation}
	j^* = \underset{j}{arg max}(\underset{k}{max} S_{j,k}).
\end{equation}

This patch selection strategy based on high similarity allows us to accurately align and distill fine-grained occlusion features, optimizing the learning effect of the model, especially in complex occlusion scenarios. We use the feature $f_{i,j^*}^{T_{adapter, patch}}$ corresponding to the highest similarity. For each target $i$, we calculate the similarity $\beta_i$ with the global feature $F_{scale}^T$:

\begin{equation}
	\beta_i={CosineSimilarity}\left(f_{i, j^*}^{T\_adapter, patch}, F_{scale}^T\right).
\end{equation}

Based on these similarity values, we define the Occlusion-Aware distillation objective:

\begin{equation}
	\mathcal{L}_{Occlusion-Aware}=\sum_{i=1}^M \beta_i \cdot\left\|F_{scale}^T-F_{scale}^S\right\|_2^2
\end{equation}

\subsection{Exponential Moving Average}

Different from traditional knowledge distillation, we do not use fixed teacher network parameters. Instead, we dynamically construct the teacher network from past iterations of the student network. In the ablation experiments described in Section 5.6, we explore different update rules for the teacher network. The experimental results show that directly copying the weights of the student network to the teacher network does not achieve model convergence. In contrast, using the exponential moving average (EMA) of the student weights, also known as the momentum encoder \cite{83}, is suitable for our framework. The update rule is as follows:

\begin{equation}
	\theta_{t} \leftarrow \tau \theta{t} + (1 - \tau) \theta_s
\end{equation}

Where $\tau$ is a decay parameter that varies from 0.996 to 1 according to a cosine schedule \cite{84}. Initially, the momentum encoder was introduced as an alternative to the queue in contrastive learning \cite{83}. However, in our framework, the role of the momentum encoder changes as we do not use a queue nor adopt a contrastive loss. By updating the teacher network using the exponential moving average (EMA), we achieve a smooth update of the teacher network by incorporating the exponentially decayed sum of historical weights. This method helps the teacher network learn more generalized feature representations that are not affected by individual image occlusion noise. In occlusion scenarios, the student network may be affected by extreme and irregular data perturbations. EMA provides a more stable learning target, allowing the student network to maintain learning continuity under dynamically changing training conditions.

\subsection{Overall Loss}

The total loss LLL used in our model training is computed as follows:

\begin{equation}
	\mathcal{L}=\mathcal{L}_{det}+\gamma_1 \mathcal{L}_{candidate}+\gamma_2 \mathcal{L}_{Occlusion-Aware},
\end{equation}

The terms $\mathcal{L}_{det}$ represent the object detection loss from Deformable DETR \cite{72}, while $\mathcal{L}_{candidate}$ and $\mathcal{L}_{Occlusion-Aware}$ are defined by equations (6) and (11) respectively. The weighting factors $\gamma_1=1$ and $\gamma_2=15$ are used to balance the contributions of the different loss components.

\section{Experiments}

\subsection{Detection Datasets}

In our experiments, we employed the recently developed large-scale outdoor orchard apple detection dataset, MinneApple \cite{87}, along with its related benchmarks. This dataset contains 1,000 images, covering over 41,000 annotated apple instances. These apples exhibit different colors and maturity stages and retain complex occlusions under natural growth conditions, without artificial removal of sparse leaves. The images were taken on both the sunny and shady sides of the tree rows, with shooting dates spanning multiple different days to ensure diversity in lighting conditions. The target instances are relatively small compared to the overall image size, and the number of targets per image can vary from 1 to 120, which meets the needs of real-world unmanned picking scenarios.

\begin{table}[ht]
	\centering
	\caption{Object Detection Accuracy Results\label{tab01}}
	\begin{tabular}{cccccc}
		\toprule
		{Method} & {AP[@0.5:0.05:0.95]} & {AP[@0.5]} & {AP[@0.75]} & {AP[small]} & {AP[large]} \\
		\midrule
		{Tiled FRCNN} & {0.341} & {0.639} & {0.339} & {0.197} & {0.208} \\
		{Faster RCNN} & {0.438} & {0.775} & {0.455} & {0.297} & {0.871} \\
		{Mask RCNN} & {0.433} & {0.763} & {0.449} & {0.295} & {0.809} \\
		{Detr} & {0.453} & {0.791} & {0.469} & {0.285} & {0.941} \\
		{Deformable Detr} & {0.512} & {0.842} & {0.520} & {0.428} & {0.943} \\
		{DetrDistill} & {0.636} & {0.894} & {0.588} & {0.479} & {0.958} \\
		{Ours} & {0.744} & {0.946} & {0.793} & {0.674} & {0.976} \\
		\bottomrule
	\end{tabular}
\end{table}

\subsection{Evaluation Metrics}

Consistent with the settings of the MinneApple \cite{87} dataset, we use Average Precision (AP) as the main evaluation metric. Specifically, we calculate the AP scores starting from an Intersection over Union (IoU) threshold of 0.5, increasing in intervals of 0.05 up to 0.95, referred to as AP@0.5:0.05:0.95. Additionally, for more detailed evaluation, we also report AP scores at IoU thresholds of 0.5 and 0.75, denoted as AP@0.5 and AP@0.75, respectively. Considering that targets of different sizes may exhibit different detection performance under natural occlusion conditions, we report the AP scores for small targets (target area less than 322 pixels) and large targets (target area greater than or equal to 922 pixels) separately.

\begin{figure}[htbp]
	\includegraphics[width=\linewidth]{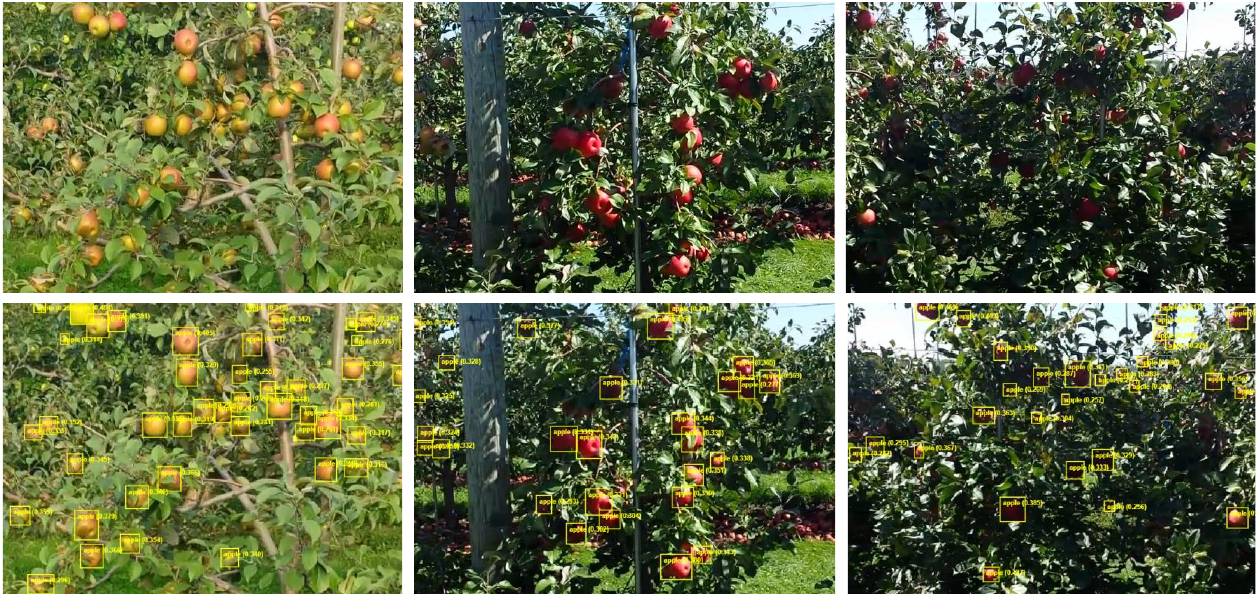}
	\caption{Qualitative Analysis: Our designed multi-scale distillation framework effectively enhances the model's object detection capabilities under various occlusion conditions, while also demonstrating good robustness to different lighting conditions. The first row of the figure shows the original images, and the second row displays the corresponding detection results.\label{fig05}}
\end{figure}

\subsection{Experimental Setup}

The experiments were conducted on four NVIDIA RTX A6000 GPUs. Unless otherwise specified, we use the pre-trained Deformable DETR \cite{72} as the backbone network for the teacher model. The student model also adopts Deformable DETR as the base framework and is optimized using the Adam optimizer \cite{88} over 50 training epochs. This experimental setup aims to explore the effect of the pre-trained backbone network in the knowledge distillation process and compare its performance.

\subsection{Quantitative Results}

As shown in Table \ref{tab01}, our method significantly outperforms traditional supervised learning-based methods in terms of bounding box detection performance. This significant performance improvement is mainly attributed to our multi-level knowledge distillation technique, which effectively learns and adapts to the impact of occlusion on target shape and semantics. Additionally, the experimental results indicate that our method demonstrates superior performance in handling small-sized target instances compared to traditional methods, successfully overcoming common challenges in small target detection.

\subsection{Qualitative Experiments}

For further validation of the effectiveness of our framework, Figure \ref{fig05} presents the comparative results of three sample images selected from the MinneApple test set. The first row of the figure displays the original test images, while the second row shows the detection results obtained using our method. These comparative images clearly demonstrate the performance of our model in detecting apples under conditions of dense occlusion and shadows. From these images, it can be observed that even in complex environments where apples are closely spaced or partially occluded, our model is able to accurately identify the apples on the trees, showcasing its excellent performance.

\subsection{Ablation Study}

To gain a deeper understanding of the specific impact of each component in the Occlusion-Enhanced Distillation (OED) framework on model performance, we report the performance of each module in detail in Table \ref{tab02}. Our baseline model, Deformable DETR without any knowledge distillation techniques applied, is labeled as Row 0 in the table, with an Average Precision (AP) of 52.6. Introducing Exponential Moving Average (EMA), occlusion enhancement, and multi-scale feature distillation techniques separately improved the model performance by 1.4 AP, 2.4 AP, and 5.9 AP, respectively. Finally, when these three techniques were applied simultaneously, the model's AP increased to 74.4, achieving a significant improvement of 13.2 AP.

\begin{table}[htbp]
	\centering
	\caption{Ablation Study\label{tab02}}
	\begin{tabular}{ccccccc}
		\toprule
		{Row} & {EMA} & {\makecell[c]{Data Occlusion \\ Enhancement}} & {\makecell[c]{Multi-scale feature \\ distillation}} & {AP} & {Ap\_small} & {AP\_large} \\
		\midrule
		{0} & {} & {} & {} & {0.512} & {0.428} & {0.943} \\
		{1} & {$\checkmark$} & {} & {} & {0.526} & {0.433} & {0.945} \\
		{2} & {} & {$\checkmark$} & {} & {0.536} & {0.454} & {0.948} \\
		{3} & {} & {} & {$\checkmark$} & {0.571} & {0.495} & {0.953} \\
		{4} & {$\checkmark$} & {$\checkmark$} & {} & {0.545} & {0.467} & {0.949} \\
		{5} & {} & {$\checkmark$} & {$\checkmark$} & {0.643} & {0.576} & {0.961} \\
		{6} & {$\checkmark$} & {$\checkmark$} & {$\checkmark$} & {0.744} & {0.674} & {0.976} \\
		\bottomrule
	\end{tabular}
\end{table}

\section{Conclusion}

In this study, we propose and thoroughly detail a novel method named "Occlusion-Enhanced Distillation" (OED), specifically designed to enhance the robustness of occluded instance detection. The OED method leverages occlusion information to normalize semantic alignment features during the learning process, thereby improving the model's capability to handle random occlusions. Specifically, we have developed a technique that integrates Grounding DINO with SAM, which can accurately extract occluded elements (such as leaves and branches) from samples and generate occlusion samples that mimic the natural growth conditions of the fruits, simulating the occlusion scenarios encountered in natural environments. Additionally, we introduced a multi-scale knowledge distillation strategy where the student network is trained with images containing occlusions, while the teacher network processes images with natural unobstructed views. This configuration allows the student network to align with the teacher network across semantic and local features at different scales, effectively narrowing the feature discrepancies between occluded and non-occluded targets. To further enhance the stability of the student network, we also employed an Exponential Moving Average (EMA) strategy, which helps the network learn feature representations that are more generalized and less affected by occlusion noise from individual images. Through a series of comprehensive comparative experiments, we demonstrate that the proposed method significantly outperforms current state-of-the-art techniques.

\section*{Declaration of interests}

The authors declare that they have no known competing financial interests or personal relationships that could have appeared to influence the work reported in this paper.

\bibliographystyle{unsrt}  
\bibliography{references}

\end{document}